\documentclass[runningheads]{llncs}
\usepackage[T1]{fontenc}
\usepackage{amsmath}
\usepackage{graphicx,verbatim}
\usepackage{booktabs}
\usepackage{pifont}
\usepackage[hyphens]{url}
\usepackage{hyperref}
\newcommand{\cross}{\ding{55}}  % cross mark
\newcommand{\tick}{\ding{51}}
\usepackage{amsmath}
\usepackage{amsfonts}   % for \mathbb
\usepackage{amssymb}    % optional       
\usepackage{tikz}
\usepackage{pgfplots}
\usepackage{pgf-pie}
\usepackage{longtable}
\usepackage[x11names]{xcolor} % or just \usepackage{xcolor}
\definecolor{SkinColor}{RGB}{0,0,255}
\definecolor{GranulationColor}{RGB}{255,0,0}
\definecolor{SloughColor}{RGB}{255,255,0}
\definecolor{MacerationColor}{RGB}{255,255,255}
\definecolor{NecroticColor}{RGB}{0,0,0}
\definecolor{BoneColor}{RGB}{255,200,124}
\definecolor{TendonColor}{RGB}{168,255,255}
\begin{document}
\title{WoundFormer: Multi-Scale Spatial Feature Fusion for Multi-Class Wound Tissue Segmentation}
\titlerunning{Multi-Class Wound Tissue Segmentation}
%% Removed for anonymized MICCAI submission
\author{Muhammad Ashad Kabir\orcidID{0000-0002-6798-6535} \and
Rabin Dulal}
\authorrunning{M. A. Kabir et al.}
\institute{School of Computing, Mathematics and Engineering,\\ Charles Sturt University, NSW, Australia\\
\email{akabir@csu.edu.au}
}

% \author{Anonymized Authors}  %% Added for anonymized MICCAI submission
% \authorrunning{Anonymized Author et al.}
% \institute{Anonymized Affiliations \\
%     \email{email@anonymized.com}}
  
\maketitle              % typeset the header of the contribution
\begin{abstract}
Chronic wounds such as diabetic foot ulcers and pressure injuries require accurate tissue-level assessment to guide treatment planning and monitor healing progression. While deep learning methods have advanced automated wound analysis, most existing approaches focus on binary segmentation and inadequately model heterogeneous tissue composition due to high intra-class variability and limited annotated data. Multi-class wound tissue segmentation, therefore, remains a challenging and clinically relevant problem.
We propose \textit{WoundFormer}, a transformer-based framework that enhances hierarchical spatial feature fusion for multi-class wound tissue segmentation. Specifically, we replace the standard SegFormer decoder with a spatially-preserving multi-scale aggregation head that maintains feature topology during cross-scale integration and strengthens contextual interactions through convolutional fusion. This design improves boundary localization and discrimination between visually similar tissue categories while preserving transformer efficiency.
We evaluate WoundFormer on the WoundTissueSeg dataset (147 images, six tissue classes) and a second benchmark (DFUTissue dataset). The proposed method achieves an overall Dice score of 81.9\%, outperforming strong CNN- and transformer-based baselines by up to 4.3 Dice points on the WoundTissueSeg benchmark, with consistent improvements across minority tissue classes. These results indicate that explicit modeling of hierarchical spatial interactions enhances transformer representations for heterogeneous wound tissue segmentation and supports more reliable quantitative wound assessment.

\keywords{Wound \and Tissue \and Segmentation  \and Deep Learning}
% Authors must provide keywords and are not allowed to remove this Keyword section.

\end{abstract}
\section{Introduction}
Chronic wounds, particularly diabetic foot ulcers and pressure injuries, affect millions of patients worldwide and pose a significant socioeconomic burden~\cite{olsson2019humanistic}. Effective wound management requires accurate assessment of tissue composition, as different tissue types reflect distinct healing states and guide treatment decisions. Granulation tissue indicates active healing, whereas necrotic tissue and slough impede recovery and often require debridement~\cite{alcantara2023identification,cutting2002maceration}. In advanced cases, exposure of deeper structures such as tendon or bone substantially increases infection risk and clinical complexity~\cite{clerici2010use}. Consequently, reliable identification and quantification of wound tissue types is critical for objective monitoring and treatment planning~\cite{bandyk2018diabetic}.

In current clinical practice, tissue assessment largely relies on visual inspection or digital planimetry~\cite{rogers2010digital}. These approaches are subjective, time-consuming, and sensitive to clinician experience. Chronic wounds often exhibit irregular geometries, poorly defined boundaries, and pronounced color heterogeneity, further complicating consistent evaluation~\cite{wang2024wound}. Such limitations hinder reproducibility and scalability, motivating automated image-based solutions.

Most existing deep learning research in wound analysis focuses on binary wound segmentation, isolating the wound region from surrounding skin~\cite{borst2025woundambit,anisuzzaman2022image}. This emphasis is partly driven by publicly available datasets and challenges, such as DFU~\cite{YAP2024103153} and FUSeg~\cite{wang2024fuseg}. In contrast, \emph{multi-class wound tissue segmentation}, which performs pixel-wise classification of heterogeneous tissue types within the wound bed, remains comparatively underexplored. This task is substantially more challenging due to high intra-class variability, visually similar tissue appearances, class imbalance, and limited annotated data. Existing approaches~\cite{RAJATHI2024105855,RW38sarp2021simultaneous,RW24niri2021superpixel,reifs2023clinical} typically address a limited number of tissue categories or evaluate on a single dataset, and thus restrict robustness.

From a modeling perspective, transformer-based segmentation frameworks have demonstrated strong global context modeling but often employ lightweight MLP decoding heads that collapse spatial structure during multi-scale feature fusion. While effective for coarse semantic segmentation, this design may limit boundary precision and fine-grained discrimination required for heterogeneous wound tissues.

To address these challenges, we propose \textit{WoundFormer}, a transformer-based framework for multi-class wound tissue segmentation that enhances hierarchical spatial feature fusion through a spatially-preserving multi-scale aggregation module. Our contributions are threefold:
\begin{itemize}
\item We introduce \textit{WoundFormer}, which replaces the standard SegFormer decoder with a spatially-aware multi-scale aggregation head to improve cross-scale contextual integration for wound tissue modeling.
\item We design a structured fusion mechanism that maintains feature map topology during decoding, strengthening boundary delineation and discrimination of visually similar tissue classes.
\item We perform extensive validation on a six-class wound tissue dataset and an additional benchmark (four-class DFUTissue dataset), demonstrating statistically significant improvements over strong CNN- and transformer-based baselines on WoundTissueSeg and competitive generalization across the additional benchmark dataset.
% across two independent datasets.
\end{itemize}

\section{WoundFormer}

\begin{figure}[!t]
    \centering
    \includegraphics[width=1.0\linewidth]{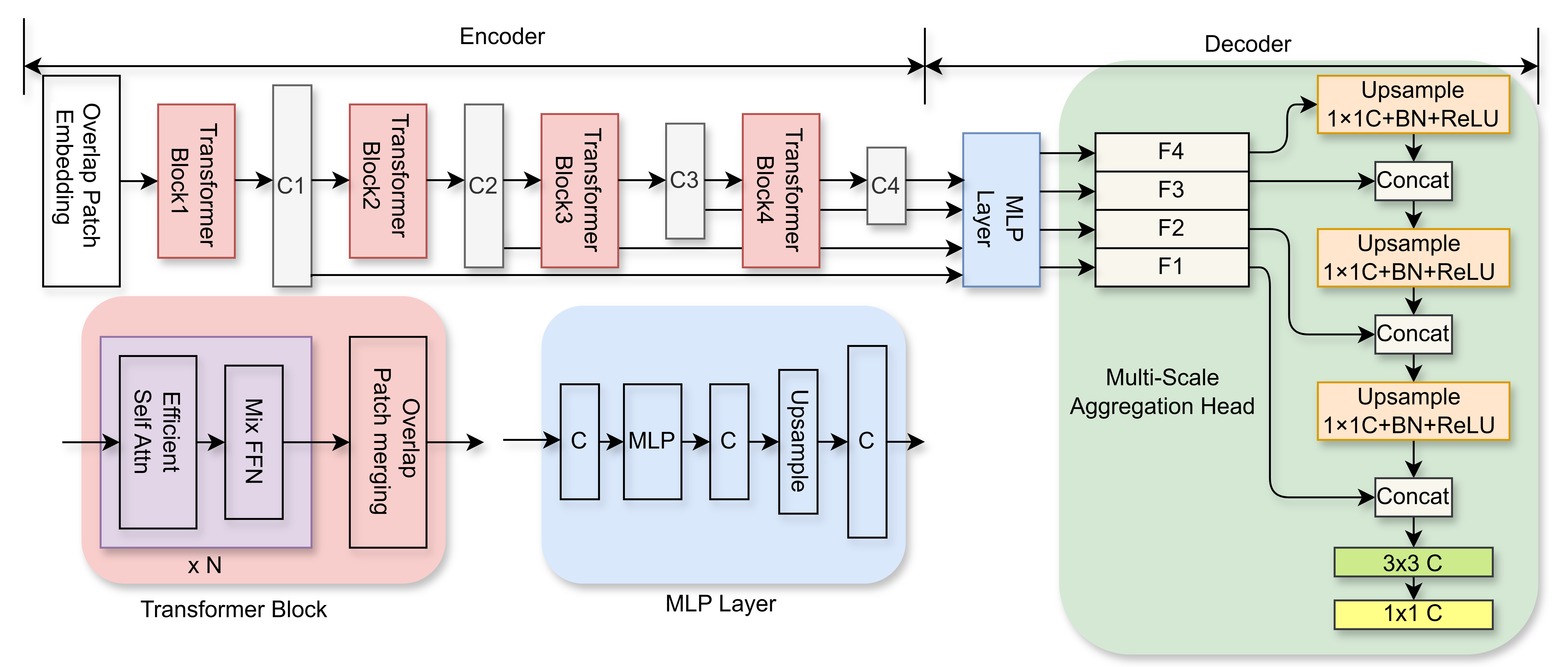}
    \caption{The proposed WoundFormer architecture consists of two main components: a hierarchical Transformer encoder, which extracts coarse and fine features, and a convolutional decoder, which fuses these multi-level features to predict the semantic segmentation mask. 
    %Here, FFN indicates a feed-forward neural network, C denotes a convolutional layer, and MLP denotes a multilayer perceptron.
    }
    \label{proposed_model}
\end{figure}

The overall architecture of WoundFormer is illustrated in Fig.~\ref{proposed_model}. 
The model consists of a hierarchical Transformer encoder and a spatially-aware multi-scale convolutional decoder.
Given an input image $I \in \mathbb{R}^{H \times W \times 3}$, the image is partitioned into overlapping $4\times4$ patches and processed by a hierarchical Transformer encoder, producing multi-scale feature maps 
$\{F_1, F_2, F_3, F_4\}$ at resolutions $\frac{1}{4}, \frac{1}{8}, \frac{1}{16}, \frac{1}{32}$ of the input, respectively. 
The decoder aggregates these features to generate a segmentation map of size $\frac{H}{4} \times \frac{W}{4} \times N_{\text{cls}}$.

\subsection{Encoder}

We adopt the Mix Transformer (MiT-B5) encoder from SegFormer~\cite{xie2021segformer}. 
The encoder architecture, including overlapped patch embedding, sequence-reduced self-attention, and Mix-FFN blocks, remains unchanged. 
Each stage outputs a feature map 
\[
F_i \in \mathbb{R}^{H_i \times W_i \times C_i}, \quad i \in \{1,2,3,4\},
\]
where lower-indexed features retain higher spatial resolution and finer structural detail.

\subsection{Spatially-Preserving Multi-Scale Decoder}

Unlike the original SegFormer All-MLP decoder, which flattens spatial tokens before fusion, we design a spatially-preserving multi-scale aggregation head that maintains feature map topology during cross-scale integration.

\paragraph{Channel Alignment.}
Each encoder feature map is projected to a unified channel dimension $C=128$ using a $1\times1$ convolution:

\[
\hat{F}_i = \sigma(\text{BN}(\text{Conv}_{1\times1}(F_i))), 
\quad \hat{F}_i \in \mathbb{R}^{H_i \times W_i \times 128},
\]
where $\sigma$ denotes ReLU activation.

\paragraph{Coarse-to-Fine Fusion.}
Fusion begins from the lowest-resolution feature:

\[
X = \hat{F}_4.
\]
At each stage $i = 3,2,1$, the fused feature $X$ is bilinearly upsampled to match the resolution of $\hat{F}_i$, concatenated along the channel dimension, and projected using a $1\times1$ convolution:

\[
X \leftarrow \sigma\big(\text{BN}(\text{Conv}_{1\times1}([\hat{F}_i, X\uparrow]))\big),
\]
where $[\cdot,\cdot]$ denotes channel concatenation and $\uparrow$ indicates bilinear upsampling. 
This hierarchical aggregation progressively integrates global semantic context with high-resolution spatial information.

\paragraph{Spatial Refinement.}
To explicitly model local spatial dependencies after fusion, we apply a $3\times3$ convolution:

\[
X_{\text{fused}} = \text{Conv}_{3\times3}(X),
\]
producing the final aggregated representation 
$X_{\text{fused}} \in \mathbb{R}^{H_1 \times W_1 \times 128}$.

\paragraph{Prediction Layer.}
A $1\times1$ convolution generates segmentation logits:

\[
M = \text{Conv}_{1\times1}(X_{\text{fused}}) 
\in \mathbb{R}^{H_1 \times W_1 \times N_{\text{cls}}}.
\]

\subsection{Design Rationale}

The proposed decoder preserves spatial structure during multi-scale fusion, mitigating the loss of local contextual information inherent to token-wise MLP decoding. 
The progressive coarse-to-fine aggregation allows high-resolution features to guide semantically rich low-resolution representations, improving boundary delineation and discrimination between visually similar tissue classes. 
By combining hierarchical transformer features with convolutional spatial refinement, the decoder strengthens local modeling while maintaining computational efficiency comparable to the original SegFormer architecture.

\section{Experiments}
\subsubsection{Datasets}
We evaluate the proposed method on two clinically acquired wound tissue datasets. 
\textbf{WoundTissueSeg}~\cite{kabir2025deep} contains 147 images with pixel-level annotations for six tissue classes: Granulation, Slough, Maceration, Necrotic tissue, Bone, and Tendon. Image resolutions vary from $128\times128$ to $2057\times2057$. We adopt the official split consisting of 118 training, 14 validation, and 15 testing images.
\textbf{DFUTissue}~\cite{dhar2024wound} comprises 110 annotated images at a fixed resolution of $256\times256$, including four tissue categories: Granulation, Callus, Fibrin, and Background. The dataset is divided into 78 training, 16 validation, and 16 testing images following the original protocol. 
Models are trained and evaluated separately on each dataset due to non-overlapping class definitions, except for the granulation class. The second dataset serves as an additional benchmark evaluation. No cross-validation was performed to maintain consistency with prior work. Class occurrence is imbalanced, and not all images contain all tissue categories.
Consistent performance improvements across both datasets support the general applicability of the proposed decoder design.

\subsubsection{Implementation Details}
All models were implemented in PyTorch using official GitHub implementations where available and initialized with released pretrained weights.
Input images were resized to $224\times224$. Training was performed using the Adam optimizer with an initial learning rate of $1\times10^{-4}$, batch size 8, and cross-entropy loss. 
A ReduceLROnPlateau scheduler (factor 0.1, patience 5) and early stopping (patience 15) were applied. For fair comparison, all models were trained and evaluated using identical dataset splits and a unified experimental protocol, including the same input resolution, optimizer, learning rate schedule, batch size, and early stopping criteria.
Performance was evaluated using the Dice similarity coefficient (DSC), with statistical significance assessed via a two-sided Wilcoxon signed-rank test.

\subsubsection{Comparison with Existing Methods}
We compare WoundFormer against widely used general-purpose segmentation architectures, including DETR~\cite{carion2020end}, Mask R-CNN~\cite{he2017mask}, HarDNet~\cite{chao2019hardnet}, SegNet~\cite{badrinarayanan2017segnet}, TransNeXt~\cite{shi2024transnext}, Deeplabv3~\cite{chen2017rethinking}, and SegFormer~\cite{xie2021segformer}, representing both CNN- and transformer-based designs. In addition, we evaluate established medical segmentation models such as nnU-Net~\cite{isensee2021nnu}, Swin UNETR~\cite{hatamizadeh2021swin}, FuSegNet~\cite{dhar2024fusegnet}, FPN+VGG16~\cite{kabir2025deep}, and DFUTissueSegNet~\cite{dhar2024wound}, which provide relevant biomedical baselines for wound analysis.

For fair comparison, all baseline models were trained without data augmentation. Under this setting, WoundFormer (without augmentation) achieves an overall Dice score of 81.9\%, outperforming all compared methods (Table~\ref{tab:comparison}). In particular, it outperforms the strongest medical baseline, FPN+VGG16 (76.5\%), by 5.4 Dice points (\textit{p} = 0.001, r = 0.76), and surpasses DFUTissueSegNet (65.0\%) by 16.9 points (\textit{p} = 0.004, r = 0.7). Among transformer-based models, SegFormer-B5 achieves 77.6\% Dice; replacing its All-MLP decoder with the proposed spatially-preserving multi-scale fusion head yields a consistent improvement of 4.3 Dice points. Performance gains are observed across most tissue categories, including challenging structures such as Bone and Tendon, where several baselines demonstrate unstable or near-zero Dice scores. These results indicate that preserving spatial structure during multi-scale feature aggregation enhances fine-grained tissue discrimination and boundary delineation in heterogeneous wound regions. Since SegFormer performed best among standard architectures, variants B0–B5 were evaluated. WoundFormer builds on the strongest backbone, improving accuracy with comparable computational cost.

Finally, incorporating data augmentation further increases performance to 85.5\% Dice, suggesting that the proposed architecture benefits from enhanced appearance variability. Augmentation strategies were adapted from DFUTissueSegNet~\cite{dhar2024wound} and include probabilistic geometric, photometric, and noise-based transformations using the same transformation parameters.

\begin{table}[!t]
    \centering
    \caption{Quantitative comparison on the WoundTissueSeg dataset. DSC (\%) is reported per class and averaged. GFLOPs (GLPs), model parameters (Param, in millions),  and inference time (seconds per image, $mean_{std}$) are provided. WoundFormer results are shown with and without data augmentation.}
    % DFUTSegNet denotes DFUTissueSegNet
       \setlength{\tabcolsep}{1pt}
    \label{tab:comparison}
    \begin{tabular}{@{\extracolsep{2pt}}l cccccccrrr}
 
    \hline
       Model  &  \multicolumn{7}{c}{DSC (\%)} & \multicolumn{2}{c}{Complexity} & Inference\\
       \cmidrule{2-8}\cmidrule{9-10}
              & Gran & Slough & Mac & Nec & Bone & Tend & Avg. & GLPs & Param & time (s)\\  
    \hline
    DETR~\cite{carion2020end}            & 55.1 & 62.3 & 45.3 & 0.0  & 0.0  & 0.0  & 27.1 & 253.0 & 60.0M & $0.85_{0.78}$\\
    Mask R-CNN~\cite{he2017mask}        & 50.3 & 31.3 & 3.5  & 15.7 & 2.0  & 0.2  & 17.1 & 134.4 & 44.4M & $0.14_{0.04}$\\
    HarDNet~\cite{chao2019hardnet}      & 1.0  & 2.4  & 1.0  & 0.7  & 0.2  & 1.2  & 1.1  & 277.8 & 51.1M & $0.01_{0.00}$\\
    SegNet~\cite{badrinarayanan2017segnet} & 72.4 & 69.7 & 11.8 & 88.0 & 4.6  & 22.9 & 44.9 & 46.7 & 31.8M & $2.39_{1.08}$\\
    TransNeXt~\cite{shi2024transnext}   & 46.8 & 61.8 & 49.1 & 0.4  & 14.6 & 0.0  & 28.8 & 476.8 & 57.7M & $0.40_{0.12}$\\
    nnU-Net~\cite{isensee2021nnu}       & 84.6 & 79.0 & 63.7 & 76.7 & 0.0 & 25.9 & 55.0 & 48.4 & 31.0M & $0.03_{0.00}$\\
    Swin UNETR~\cite{hatamizadeh2021swin} & 73.1 & 0.8  & 45.3 & 0.0  & 0.0 & 2.6  & 20.3 & 58.7 & 44.9M & $0.04_{0.00}$\\
    FuSegNet~\cite{dhar2024fusegnet}    & 58.2 & 65.6 & 18.8 & 70.7 & 0.1 & 21.9 & 39.1 & 71.0 & 70.9M & $0.07_{0.01}$\\
    Deeplabv3~\cite{chen2017rethinking} & 56.8 & 40.1 & 14.8 & 54.4 & 30.4 & 21.4 & 36.3 & 178.7 & 42.0M & $0.20_{0.33}$\\
    FPN+VGG16~\cite{kabir2025deep}    & 89.6 & 72.5 & 82.8 & 78.3 & 67.2 & 66.8 & 76.5 & 21.9 & 17.6M & $0.29_{0.76}$\\
    SegFormer-B0~\cite{xie2021segformer} & 71.6 & 83.6 & 59.6 & 98.6 & 45.4 & 54.9 & 64.6 & 8.4 & 3.9M & $0.03_{0.03}$\\
    SegFormer-B1~\cite{xie2021segformer} & 70.3 & 83.8 & 72.3 & 81.3 & 58.2 & 58.3 & 70.7 & 15.9 &13.7M & $0.06_{0.03}$\\
    SegFormer-B2~\cite{xie2021segformer} & 93.0 & 60.2 & 74.7 & 88.5 & 64.1 & 67.2 & 74.6 & 62.4 & 27.5M & $0.19_{0.10}$\\
    SegFormer-B3~\cite{xie2021segformer} & 79.9 & 83.2 & 70.8 & 86.9 & 46.0 & 67.1 & 72.3 & 79.0 & 47.3M & $0.26_{0.13}$\\
    SegFormer-B4~\cite{xie2021segformer} & 64.7 & 79.9 & 76.3 & 0.5  & 55.8 & 15.8 & 48.8 & 95.7 & 64.1M & $0.32_{0.17}$\\
    SegFormer-B5~\cite{xie2021segformer} & 91.4 & 73.3 & 81.0 & 82.7 & 76.6 & 60.5 & 77.6 & 183.3 & 84.6M & $2.15_{1.09}$\\
    DFUTissueSegNet~\cite{dhar2024wound} & 89.4 & 63.2 & 82.6 & 86.1 & 68.2 & 0.4 & 65.0 & 7.8 & 51.1M & $2.02_{1.61}$\\
    \hline
    WoundFormer      & 87.3 & 92.0 & 81.0 & 83.6 & 71.7 & 75.8 & \textbf{81.9} & 183.6 & 84.9M & $2.08_{0.69}$\\
 
    % DFUTSegNet~\cite{dhar2024wound}  & 91.0 & 66.1 & 85.2 & 88.2 & 76.2 & 1.4 & 67.9 & 7.8  & 51.1 & $2.02_{1.61}$\\
    (w/o augmentation)\\
    WoundFormer      & 87.1 & 89.1 & 92.2 & 89.3 & 77.1 & 78.2 & \textbf{85.5} & 183.6 & 84.9M & $2.08_{0.69}$\\
    (w augmentation)\\
    \hline
    \end{tabular}
\end{table}

\subsubsection{Qualitative Comparison}
Based on the top five performing models in Table~\ref{tab:comparison}, qualitative segmentation results on representative wound samples are shown in Fig.~\ref{fig:wound_comparison}. Compared with competing methods, WoundFormer demonstrates improved boundary delineation and more consistent tissue classification, particularly in regions with complex morphology and heterogeneous appearance. In several examples, competing models exhibit over-smoothing or misclassification between visually similar tissue types, whereas WoundFormer better preserves fine-grained structures and reduces spurious predictions. The predicted masks show closer spatial alignment with ground-truth annotations, especially for small or irregular tissue regions. Notably, improvements are evident in minority classes such as Bone and Tendon, where baseline methods often under-segment or fail to detect these structures.

\newlength{\imgsizew}
\setlength{\imgsizew}{1.6cm}
\newlength{\imgsizeh}
\setlength{\imgsizeh}{1.5cm}
\begin{figure*}[!t]
\centering
\footnotesize
\setlength{\tabcolsep}{1pt}
\renewcommand{\arraystretch}{0.5}
\scriptsize
% =======================
% Table of images
% =======================
\begin{tabular}{cccccccc}

 & Input & GT & nnU-Net & FPN+VGG16 & SegFormerB5 & DFUTissueSeg & WoundFormer \\

\rotatebox{90}{Sample 1} 
& \includegraphics[width=\imgsizew,height=\imgsizeh]{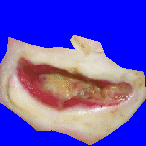}
& \includegraphics[width=\imgsizew,height=\imgsizeh]{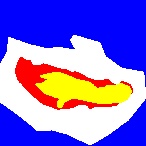}
& \includegraphics[width=\imgsizew,height=\imgsizeh]{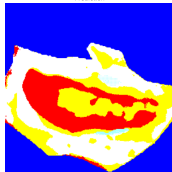}
& \includegraphics[width=\imgsizew,height=\imgsizeh]{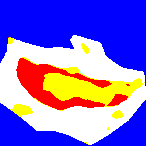}
& \includegraphics[width=\imgsizew,height=\imgsizeh]{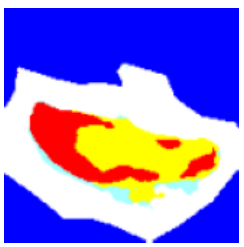}
& \includegraphics[width=\imgsizew,height=\imgsizeh]{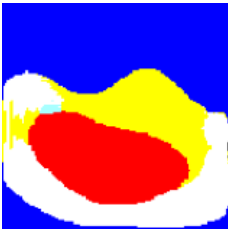}
& \includegraphics[width=\imgsizew,height=\imgsizeh]{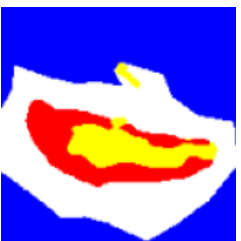}
\\

\rotatebox{90}{Sample 2} 
& \includegraphics[width=\imgsizew,height=\imgsizeh]{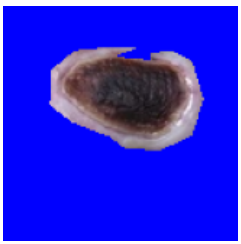}
& \includegraphics[width=\imgsizew,height=\imgsizeh]{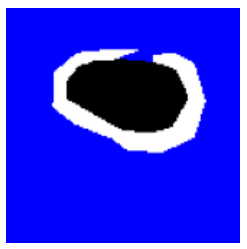}
& \includegraphics[width=\imgsizew,height=\imgsizeh]{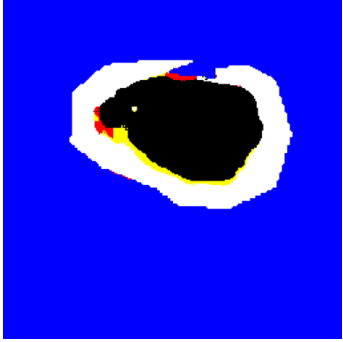}
& \includegraphics[width=\imgsizew,height=\imgsizeh]{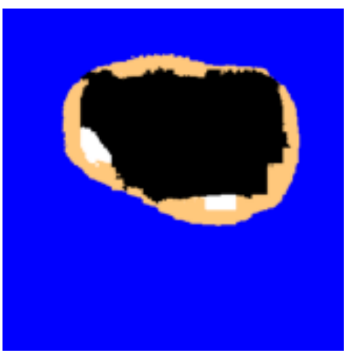}
& \includegraphics[width=\imgsizew,height=\imgsizeh]{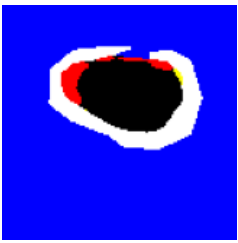}
& \includegraphics[width=\imgsizew,height=\imgsizeh]{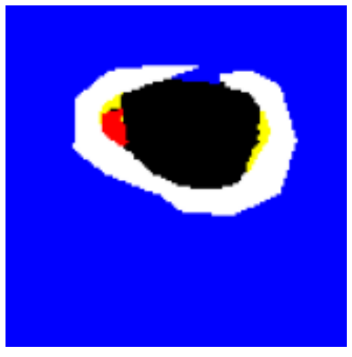}
& \includegraphics[width=\imgsizew,height=\imgsizeh]{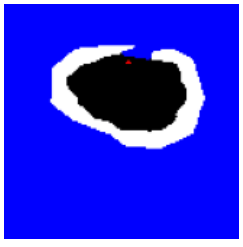}
\\

\rotatebox{90}{Sample 3} 
& \includegraphics[width=\imgsizew,height=\imgsizeh]{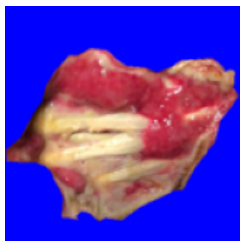}
& \includegraphics[width=\imgsizew,height=\imgsizeh]{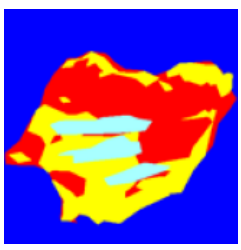}
& \includegraphics[width=\imgsizew,height=\imgsizeh]{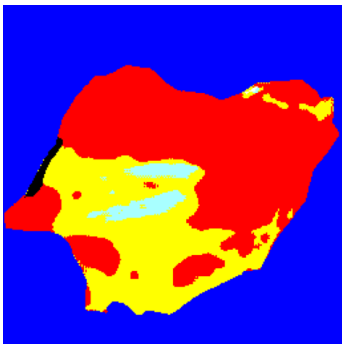}
& \includegraphics[width=\imgsizew,height=\imgsizeh]{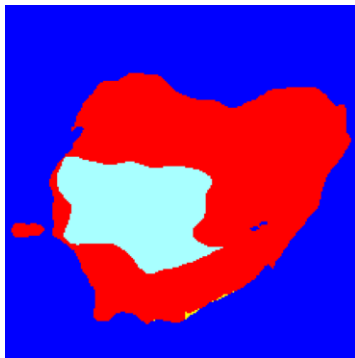}
& \includegraphics[width=\imgsizew,height=\imgsizeh]{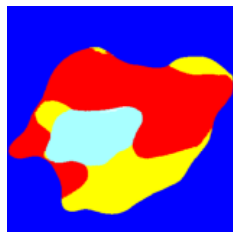}
& \includegraphics[width=\imgsizew,height=\imgsizeh]{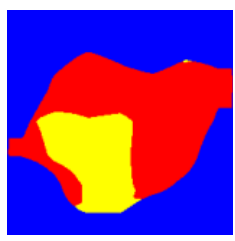}
& \includegraphics[width=\imgsizew,height=\imgsizeh]{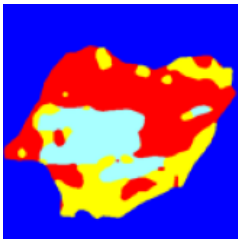}
\\

\rotatebox{90}{Sample 4} 
& \includegraphics[width=\imgsizew,height=\imgsizeh]{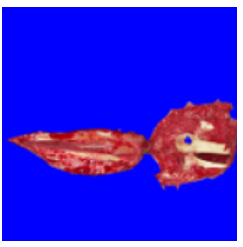}
& \includegraphics[width=\imgsizew,height=\imgsizeh]{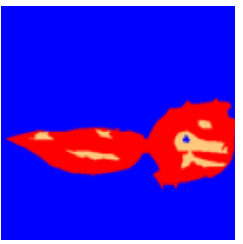}
& \includegraphics[width=\imgsizew,height=\imgsizeh]{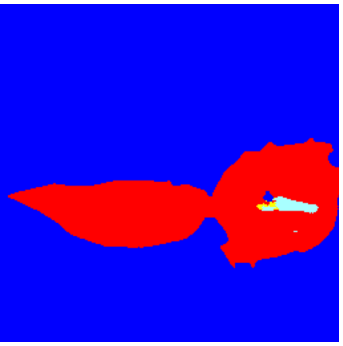}
& \includegraphics[width=\imgsizew,height=\imgsizeh]{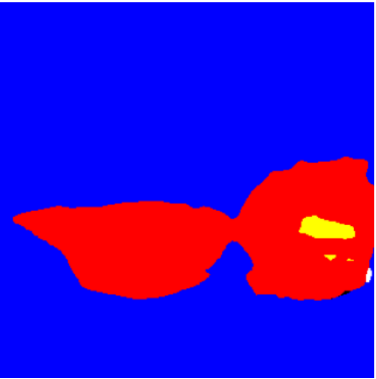}
& \includegraphics[width=\imgsizew,height=\imgsizeh]{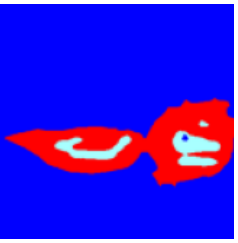}
& \includegraphics[width=\imgsizew,height=\imgsizeh]{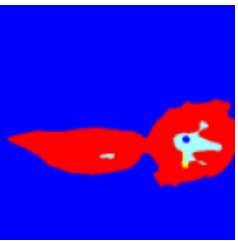}
& \includegraphics[width=\imgsizew,height=\imgsizeh]{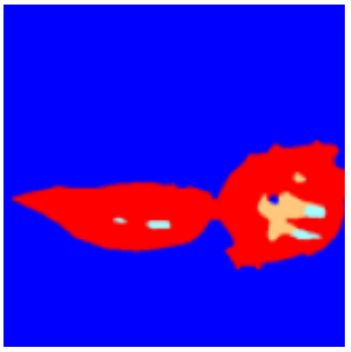}
\\

\end{tabular}

% =======================
% Legend of tissue types (exact RGB, black border)
% =======================
\begin{tabular}{cccccc}
% \fcolorbox{black}{SkinColor}{\color{SkinColor}\rule{12pt}{12pt}} Mask Back &
\fcolorbox{black}{GranulationColor}{\color{GranulationColor}\rule{12pt}{12pt}} Granulation &
\fcolorbox{black}{SloughColor}{\color{SloughColor}\rule{12pt}{12pt}} Slough &
\fcolorbox{black}{MacerationColor}{\color{MacerationColor}\rule{12pt}{12pt}} Maceration &
\fcolorbox{black}{NecroticColor}{\color{NecroticColor}\rule{12pt}{12pt}} Necrotic &
\fcolorbox{black}{BoneColor}{\color{BoneColor}\rule{12pt}{12pt}} Bone &
\fcolorbox{black}{TendonColor}{\color{TendonColor}\rule{12pt}{12pt}} Tendon
\end{tabular}

% =======================
% Caption
% =======================
\caption{Qualitative comparison of segmentation results on representative samples from the WoundTissueSeg dataset. From left to right: input, ground truth, nnU-Net, FPN+VGG16, SegFormer-B5, DFUTissueSegNet, and WoundFormer. Color legend indicates tissue classes.}
\label{fig:wound_comparison}
\end{figure*}

\begin{table}[!t]
    \centering
    \caption{Abalation analysis. A. Conv. means additional convolutional layers.}
    \label{tab:placeholder}
    
    \begin{tabular}{cccccc|c}
    \hline
      Conv. & Batch Norm & Activation & A. Conv. & Loss Function & Augmentation & DSC \\
      \hline
      
    $1\times1$ & \cross  & \cross   &   --    & Cross Entropy & \cross  & 51.11 \\
    $1\times1$ & \tick & \cross   &   --    & Cross Entropy & \cross  & 57.18 \\
    $1\times1$ & \tick & GeLU &   --    & Cross Entropy & \cross  & 66.13 \\
    $1\times1$ & \tick & ReLU &   --    & Cross Entropy & \cross  & 67.67 \\
    $3\times3$ & \tick & ReLU &   --    & Cross Entropy & \cross  & 65.01 \\
    $1\times1$ & \tick & ReLU & $1\times1$ & Cross Entropy & \cross  & 67.69 \\
    $1\times1$ & \tick & ReLU & $1\times1$, $1\times1$ & Cross Entropy & \cross  & 46.50 \\
    $1\times1$ & \tick & ReLU & $3\times3$ & Cross Entropy & \cross  & 53.28 \\
    $1\times1$ & \tick & ReLU & $1\times1$, $3\times3$ & Cross Entropy & \cross  & 81.89 \\
    $1\times1$ & \tick & ReLU & $1\times1$, $3\times3$ & Focal+Dice & \tick & 45.78 \\
    $1\times1$ & \tick & ReLU & $1\times1$, $3\times3$ & Cross Entropy & \tick & 85.50 \\
    \hline
    \end{tabular}
    
\end{table}
\subsubsection{Ablation Analysis}
Table~\ref{tab:placeholder} presents an ablation study evaluating the impact of key decoder components. 
Starting from a baseline $1\times1$ convolution without normalization or activation (51.11\% DSC), adding Batch Normalization and ReLU improves performance to 67.67\%, indicating the importance of feature normalization and non-linearity in stabilizing multi-scale fusion. 
Replacing the $1\times1$ convolution with a single $3\times3$ layer does not yield additional improvement, suggesting that increased kernel size alone is insufficient. In contrast, introducing stacked convolutional layers combining $1\times1$ and $3\times3$ operations increases performance to 81.89\%, demonstrating the benefit of hierarchical spatial refinement within the decoder.
Comparing loss functions shows that cross-entropy outperforms Focal+Dice in this setting. Finally, incorporating data augmentation further improves performance to 85.50\%, indicating that the proposed decoder benefits from enhanced appearance variability during training.

\subsubsection{Independent Benchmark Evaluation}
Table~\ref{tab:dfutissue_dsc} reports evaluation results of the DFUTissue dataset. 
When trained and evaluated using the official split, WoundFormer achieves 78.34\% Dice without augmentation versus 79.49\% for DFUTissueSegNet, with a statistically significant difference (\textit{p} = 0.025, r = 0.50). 
Compared to general-purpose segmentation architectures, WoundFormer demonstrates improved and more balanced performance across tissue classes, particularly for Granulation and Callus. With data augmentation, performance increases to 85.40\%, approaching the best-performing model (87.08\%) with no significant difference (\textit{p} = 0.56, r = 0.15) and consistent class-wise accuracy. These findings indicate that the proposed spatially-preserving decoder design remains effective across multiple wound tissue segmentation datasets with differing tissue taxonomies.
% independent clinical cohorts and varying tissue taxonomies.
\begin{table}[!t]
\centering
\caption{Quantitative comparison on the DFUTissue dataset. 
% Dice similarity coefficient (DSC, \%) is reported per class (Granulation, Fibrin, Callus, Background) and averaged across classes. Inference time (seconds per image, mean$\pm$SD) is also provided. 
% Results are shown with and without data augmentation where applicable.
}
\label{tab:dfutissue_dsc}
\begin{tabular}{@{\extracolsep{4pt}}l ccccc r}
\hline
Model  &  \multicolumn{5}{c}{DSC (\%)} & Inference\\
\cmidrule{2-6}
       & Gran & Fibrin & Callus & Back & Avg. & time (s)\\  
\hline

DETR~\cite{carion2020end}       & 64.54 & 0.00  & 31.47 & 87.61 & 45.91 & 0.03$\pm$0.01\\
Mask R-CNN~\cite{he2017mask}        & 88.64 & 7.72  & 0.00  & 97.79 & 48.54 & 0.76$\pm$0.14\\
HarDNet68~\cite{chao2019hardnet}         & 85.11 & 11.10 & 57.43 & 89.27 & 60.73 & 0.02$\pm$0.00\\
SegNet~\cite{badrinarayanan2017segnet}            & 21.44 & 7.39  & 0.00  & 91.90 & 30.18 & 2.29$\pm$0.67\\
TransNeXt~\cite{shi2024transnext}         & 70.32 & 6.36  & 0.00  & 93.94 & 42.66 & 1.44$\pm$0.18\\
nnU-Net~\cite{isensee2021nnu}           & 25.56 & 0.00  & 0.00  & 88.22 & 28.46 & 0.03$\pm$0.00\\
Swin UNETR~\cite{hatamizadeh2021swin}        & 89.68 & 3.81  & 0.00  & 89.15 & 45.66 & 0.79$\pm$0.27\\
FuSegNet~\cite{dhar2024fusegnet}          & 83.43 & 11.64 & 0.00  & 85.49 & 45.14 & 0.06$\pm$0.01\\
Deeplabv3~\cite{chen2017rethinking}         & 86.20 & 12.50 & 48.10 & 90.30 & 59.28 & 0.09$\pm$0.07\\
FPN+VGG16~\cite{kabir2025deep}       & 80.30 & 20.40 & 55.10 & 88.60 & 61.10 & 0.13$\pm$0.00\\
SegFormer-B0~\cite{xie2021segformer}      & 89.32 & 2.91  & 42.52 & 89.21 & 56.01 & 0.09$\pm$0.07\\
SegFormer-B5~\cite{xie2021segformer}      & 88.20 & 62.35 & 74.90 & 78.83 & 76.07 & 0.13$\pm$0.00\\
DFUTissueSegNet~\cite{dhar2024wound} & 90.75 & 68.40 & 82.31 & 76.50 & 79.49 & 0.57$\pm$0.12 \\
WoundFormer & 82.45 & 69.32 & 86.71 & 74.88 & 78.34 & 1.07$\pm$0.09\\
\hline
DFUTissueSegNet (w augmentation)   & 91.99 & 74.52 & 85.91 & 95.90 & 87.08 & 0.57$\pm$0.12\\
WoundFormer (w augmentation)       & 88.27 & 70.75 & \textbf{87.42} & 95.13 & 85.40 & 1.07$\pm$0.09\\

\hline
\end{tabular}
\end{table}

\section{Conclusions}
We presented WoundFormer, a transformer-based framework for multi-class wound tissue segmentation with a spatially-preserving multi-scale decoder. By maintaining feature map topology during hierarchical fusion, the proposed design improves boundary delineation and fine-grained tissue discrimination. 
Experimental results on two clinical datasets demonstrate that the proposed decoder achieves competitive performance for multi-class wound tissue segmentation, while maintaining computational efficiency.
% Experimental results on two clinical datasets demonstrate consistent performance gains over strong CNN- and transformer-based baselines. 
These findings suggest that explicitly modeling spatial interactions in transformer decoders is beneficial for heterogeneous wound tissue segmentation. Future work will explore integration with automated clinical reporting and larger multi-center validation.
\bibliographystyle{splncs04}
\bibliography{reference}

%\end{thebibliography}
\end{document}